\documentclass{article}

\PassOptionsToPackage{numbers, compress}{natbib}

\usepackage[main, final]{neurips_2026}

\usepackage[utf8]{inputenc} 
\usepackage[T1]{fontenc}    
\usepackage{hyperref}       
\usepackage{url}            
\usepackage{booktabs}       
\usepackage{amsfonts}       
\usepackage{nicefrac}       
\usepackage{microtype}      
\usepackage{xcolor}         

\usepackage{graphicx}
\usepackage{amsmath}
\usepackage{multirow}
\usepackage{enumitem}

\usepackage[normalem]{ulem}
\usepackage{wrapfig}

\title{REINS: Learning Inertia-Induced Geometry for Physics-Consistent Motion Representation in Clinical Gait Phenotyping}

\author{%
  Yiran Ding$^{1}$\thanks{This work is done when the author is a research intern at MSU.} \quad
  Yufei Zhang$^{2}$ \quad
  Zijun Cui$^{1}$ \\
  $^{1}$Michigan State University\\
  $^{2}$Independent Researcher \\
  \texttt{yiran\_ding@alumni.brown.edu}, \quad
  \texttt{yufeizhang96@outlook.com}, \quad
  \texttt{cuizijun@msu.edu}
}

\begin{document}

\maketitle

\begin{abstract}
  Parkinson's disease and other gait-related neurological disorders manifest in subtle but biomechanically meaningful changes in how patients move, making computational gait phenotyping a valuable tool for clinical assessment. Yet existing latent representations of motion are typically learned from data alone and miss the biomechanical structure that gives clinical motion its diagnostic meaning, leaving them vulnerable when training cohorts are small or acquisition sites differ. We propose \textbf{REINS} (\textbf{R}iemannian \textbf{E}mbedding for \textbf{IN}ertia-induced Motion \textbf{S}paces), an Euler-Lagrange informed motion representation that uses the generalized inertia matrix of articulated systems to define a biomechanically grounded geometry on configuration space. Because kinetic energy is itself a quadratic form in velocity weighted by the inertia matrix, this metric is a mechanically determined Riemannian structure, and REINS learns a latent embedding of the resulting structured space. By defining the representation space itself through rigid-body dynamics rather than enforcing physics as a post-hoc penalty, REINS aligns the geometry of motion representation with the same physical laws clinicians rely on to interpret gait. On Parkinson's disease (PD) gait analysis, REINS measures deviation from healthy motion as a Riemannian distance under the inertia-induced metric, and improves healthy-control (HC) vs.\ PD discrimination and Unified Parkinson's Disease Rating Scale (UPDRS) gait severity correlation. Code is available at \url{https://github.com/DominoAI-Lab/REINS-NeurIPS-2026}.

\end{abstract}

\section{Introduction}

Human gait and movement carry rich diagnostic information for a range of neurological disorders, including Parkinson's disease (PD), where motor symptoms such as bradykinesia, reduced stride length, and increased stride variability are central to clinical assessment~\cite{hausdorff1998gait,hausdorff2009gait}. Recent advances in monocular 3D human pose and motion estimation~\cite{zhang2023body,shin2024wham,zhang2024physpt} have made it feasible to assess these behaviors automatically from video, opening the door to scalable, non-intrusive clinical evaluation.

Despite this potential, clinical motion datasets remain limited in scale, cohort diversity, and label quality, which makes purely data-driven motion representations vulnerable to overfitting and poor cross-cohort generalization \cite{adeli2025care}. Human gait, however, is the motion of an articulated rigid-body system, a biomechanical structure that holds across subjects and sites. Rigid-body dynamics therefore provide a universal, dataset-independent prior that constrains motion representations regardless of how much clinical data is available.

Existing physics-informed motion methods do incorporate physical laws, but typically as a post-hoc trajectory-level regularizer for reconstruction quality \cite{shimada2020physcap,yuan2021simpoe,zhang2024physpt,zhang2024incorporating}, without changing the geometry of the representation space itself. A separate line of work in robotics and motion learning has long recognized that the inertia matrix induces a Riemannian metric on configuration space and exploited this geometry for planning, control, and skill learning \cite{jaquier2022riemannian,li2024riemannian,beik2021learning}. Yet using inertia-induced geometry to shape a learned representation of human motion for clinical phenotyping remains largely unexplored. Such a representation is well-suited to clinical phenotyping, where phenotype scores are expected to be grounded in shared mechanical principles, to transfer across acquisition sites and patient cohorts \cite{adeli2025care}, to support subject-level severity tracking when UPDRS ratings are sparse or unavailable \cite{guan2025zero}, and to be obtained from monocular video alone, without requiring specialized motion-capture facilities \cite{shin2024wham}.

\begin{figure}[!t]
\centering
\includegraphics[width=\textwidth]{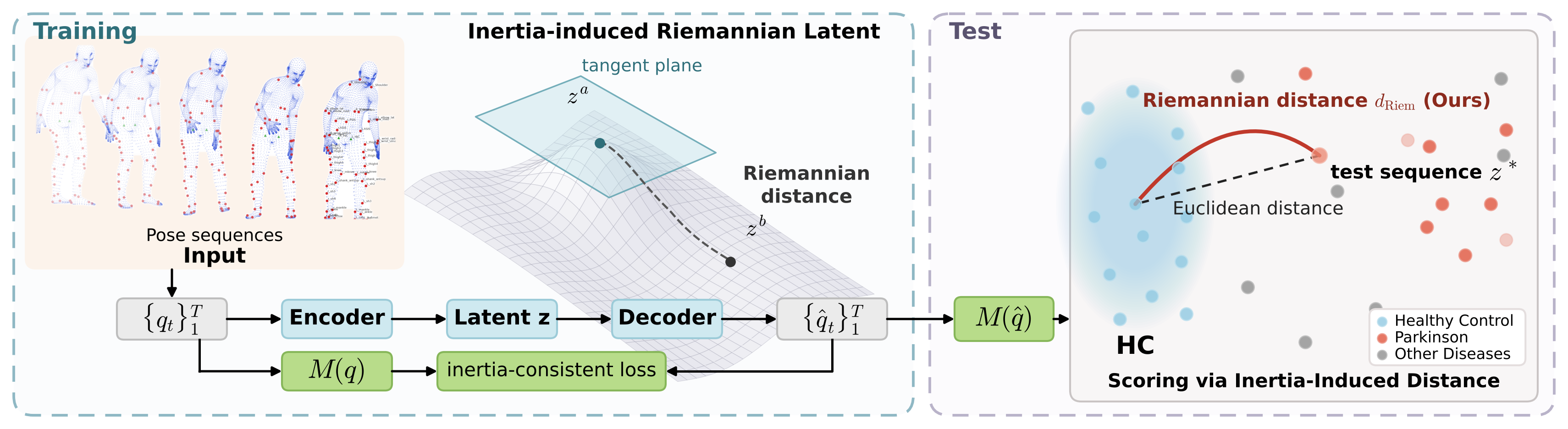}
\caption{Overview of REINS. Training (left): a pose sequence is encoded into a latent vector and reconstructed by the decoder. The inertia-consistent loss weights the reconstruction velocity error by the body's inertia matrix $M(q)$, so that the latent space inherits the inertia-induced geometry (surface). Distances between embedded sequences $z^a$ and $z^b$ are measured by the Riemannian distance under this geometry. Testing (right): a test sequence is embedded as $z^*$. The inertia matrix $M(\hat q)$ of its decoded pose defines the local metric, and the sequence is scored by its Riemannian distance $d_{\text{Riem}}$ to the healthy-control (HC) reference embeddings. The score separates subjects with Parkinson's disease from healthy controls and correlates with clinical severity on cohorts not used for training. The Riemannian distance (ours) separates these groups more reliably than Euclidean distance (dashed).} 
\label{fig:pipeline}
\end{figure}

In this work, we study physics-informed motion representations through a geometric lens. Because the kinetic energy of a rigid-body system takes the form \(\mathcal{T}(q,\dot q)=\tfrac{1}{2}\dot q^\top M(q) \dot q\), the inertia matrix \(M(q)\) is a mechanically determined Riemannian metric on configuration space. We use this inertia-induced geometry as the foundation of representation, rather than enforcing Euler-Lagrange consistency as a post-hoc trajectory constraint. Figure~\ref{fig:pipeline} provides an overview of REINS. Our contributions are:
\begin{itemize}[leftmargin=1.5em, itemsep=2pt, topsep=2pt]
    \item \textbf{A Motion Representation for Parkinson's Gait Phenotyping.} We frame computational gait phenotyping as representation geometry rather than reconstruction quality, and motivate biomechanics as the inductive bias that aligns representation structure with how clinicians read gait.
    \item \textbf{A Geometric Lens of Biomechanics Integration.} We introduce REINS, in which the mechanically determined inertia metric, computed from the body model and pose, structures both the learned motion representation through the inertia-consistent loss and its Riemannian-distance scoring for clinical gait phenotyping.
    \item \textbf{Empirical Evaluation on Clinical PD Gait Phenotyping.} We evaluate REINS on the CARE-PD~\cite{adeli2025care} benchmark across full-data, reduced-data, and cross-cohort transfer regimes, on healthy-control (HC) vs.\ PD discrimination and UPDRS-gait severity correlation, with a factorial ablation that separates the inertia-consistent loss from the scoring function.
\end{itemize}

\section{Related Works}
\noindent\textbf{Computational Gait Analysis for Parkinson's Disease.}
Automated gait assessment for PD has been studied with various sensing modalities, including inertial measurement units (IMUs) and video cameras~\cite{lu2020vision,endo2022gaitforemer}. Among these modalities, vision-based methods are non-intrusive, scalable, and more relevant to our setting. They typically extract 2D or 3D pose from clinical recordings and apply supervised classifiers or regressors to predict UPDRS-gait scores~\cite{lu2020vision}. To address label scarcity, some methods adopt self-supervised pretraining~\cite{endo2022gaitforemer}, while zero-shot approaches use generative models of healthy motion to flag pathological deviation without severity labels~\cite{guan2025zero}. Beyond single-cohort evaluation, recent work increasingly emphasizes generalization across clinical sites~\cite{adeli2025care}. Nonetheless, existing methods treat motion primarily as a kinematic signal and pursue generalization through architectural and pretraining choices. How mechanical structure, in particular the inertia geometry of rigid-body dynamics, can shape motion representations under clinical data scarcity and cross-site shift remains largely unexamined.

\noindent\textbf{Physics-Informed Human Motion Modeling.}
Physics provides a strong prior for making estimated, predicted, or generated motion physically plausible. One line of work relies on a physics simulator to incorporate physics, either driving a simulated character to reproduce a target motion with a learned control or imitation policy~\cite{yuan2021simpoe,yuan2023physdiff}, or refining noisy motion estimates through post hoc trajectory optimization~\cite{shimada2020physcap,gartner2022trajectory}. Another line of work instead integrates physical principles directly into learning models. For example, PhysMoP integrates the rigid-body dynamics equations through neural network design, significantly reducing the error of motion prediction from past observations~\cite{zhang2024incorporating}. PhysPT explicitly models joint torques and contact forces and introduces an Euler-Lagrange residual loss during training to produce physically plausible motion estimates from monocular video~\cite{zhang2024physpt}. In both cases, the goal is to improve the physical plausibility of motion. They primarily enforce physics as a consistency or regularization constraint at the trajectory level, while only modeling healthy adults. In contrast, we take the latent representation itself as the primary object of study, and ask how the inertia-induced geometry of rigid-body dynamics can structure that representation for clinical motion analysis.

\noindent\textbf{Geometry in Motion Representation Learning.}
Learning latent representations of motion has been widely studied. For human motion, existing works primarily build on established frameworks whose latent spaces have no mechanical metric~\cite{pavlakos2019expressive}, such as conditional variational autoencoders that model per-frame transitions~\cite{rempe2021humor} or entire sequences~\cite{petrovich2021action}. For other dynamical systems, some works constrain the latent states with known dynamics equations for reconstruction and physical parameter estimation~\cite{yang2022learning}. However, these works do not study how distances between latent states should be measured. In robotics, Riemannian metrics derived from the inertia matrix~\cite{jaquier2022riemannian,li2024riemannian} or learned from demonstrations~\cite{beik2021learning} are used for motion learning, control, and planning, but they serve to generate robot motions rather than to compare motions in a learned representation. For learned representations, pullback geometry transfers a data-space metric to the latent space through the decoder~\cite{arvanitidis2017latent,arvanitidis2020geometrically}, but this metric is typically Euclidean, learned, or hand-designed, leaving open which metric suits human motion. To our knowledge, REINS is the first to equip a learned human motion representation with the inertia-induced metric, so that deviation from healthy gait is measured by mechanical effort rather than kinematic extent.

\section{Methodology}


We consider the task of learning a low-dimensional representation of human motion that is structured by physics rather than by data alone. Given a pose sequence $q_{1:T}$ describing the configuration of an articulated human body across $T$ frames, we seek an encoder $\phi$ producing a latent $z = \phi(q_{1:T})$ such that distances in latent space reflect physically meaningful deviation between motions. Instead of using physics to refine reconstructed motion as a post-processing step, we use it to shape the geometry of the representation space itself. Section~\ref{sec:prelim} reviews the rigid-body dynamics and explains why the inertia matrix defines a Riemannian metric. Section~\ref{sec:motion_space} defines the inertia-induced motion space. In Section~\ref{sec:learning_latent}, we learn a latent embedding of this space with an inertia-consistent reconstruction loss. Section~\ref{sec:scoring} defines the score using Riemannian distance in latent space and describes how we use the score for downstream tasks.

\subsection{Mechanical Preliminaries}
\label{sec:prelim}

\noindent\textbf{Rigid-Body Dynamics.}
We treat the human body as an articulated rigid-body system with generalized coordinates $q \in \mathbb{R}^n$ describing root translation and joint angles. Its dynamics are governed by the Euler-Lagrange equation:
\begin{equation}
M(q)\ddot{q} + C(q,\dot{q})\dot{q} + g(q) = \tau + J_C^\top(q)\lambda,
\end{equation}
where $M(q) \in \mathbb{R}^{n \times n}$ is the generalized inertia matrix, $C(q,\dot q)\dot q$ collects Coriolis and centrifugal forces, $g(q)$ is the gravitational force, $\tau$ are the joint torques produced by muscles, and $J_C^\top(q)\lambda$ is the contact wrench from the environment. The inertia matrix $M(q)$ is the dynamic counterpart of the body's mass distribution at configuration $q$, since each entry $M_{ij}(q)$ encodes how much of the joint torque $\tau_i$ is required to produce a unit acceleration in coordinate $j$, with $M$ aggregating contributions from all body segments through the kinematic chain.

\noindent\textbf{Inertia as a Riemannian Metric.}
The same matrix defines the kinetic energy of the system,
\begin{equation}
\mathcal{T}(q,\dot q)=\tfrac{1}{2}\dot q^\top M(q)\dot q.
\end{equation}
Because $M(q)$ is symmetric and positive definite for any physically realizable body configuration, it defines a Riemannian metric on the configuration space~\citep{jaquier2022riemannian}. Geometrically, the unit ball of this metric stretches along directions in which the body has low effective inertia and contracts along directions in which it has high effective inertia, so distances under $M(q)$ are induced by the kinetic-energy metric and measure the mechanical effort of a motion rather than its kinematic extent. In gait, this geometry directly reflects the way a clinician interprets movement: changes in stride length, posture, or limb swing alter the body's effective inertia and therefore the mechanical work required to produce the same kinematic trajectory. At the same angular speed, hip motion has about four times the kinetic energy of shoulder motion and about 260 times that of wrist motion. Euclidean joint-angle distance weights all three equally. Posture also affects the metric: flexing the knee to $90^\circ$ lowers the effective hip inertia from 2.0 to 1.2\,kg\,m$^2$.

\noindent\textbf{Choice of Inertia Matrix.}
We use $M(q)$ to define the geometry, rather than torques or muscle quantities, for two reasons. First, this geometry follows directly from rigid-body dynamics. The body is an articulated rigid-body system governed by the Euler-Lagrange equation above. Its kinetic energy is $\tfrac{1}{2}\dot q^\top M(q)\dot q$, where $M(q)$ defines a symmetric positive-definite form on velocities and hence a Riemannian metric~\cite{arnold1989mathematical}. Second, $M(q)$ can be computed in closed form from the pose and body model. In contrast, computing torques requires contact wrenches, which can be measured with force plates or estimated with a learned solver~\cite{liu2025imdy}. Musculoskeletal models driven by surface EMG~\cite{zhang2022physics} require signals that CARE-PD does not provide. 

\subsection{Defining the Inertia-Induced Motion Space}

\label{sec:motion_space}

\begin{wrapfigure}{r}{0.6\linewidth}
    \vspace{-\baselineskip}
    \centering
    \includegraphics[width=0.85\linewidth]{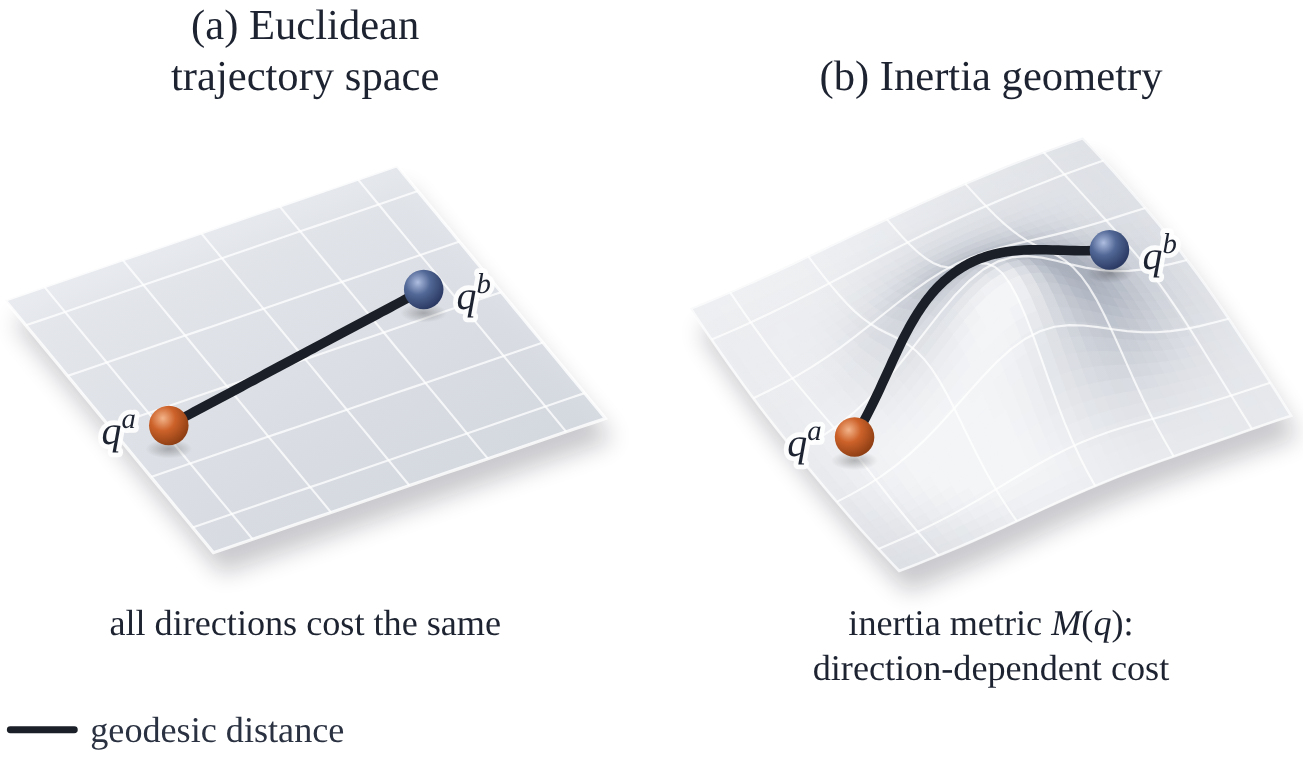}
    \caption{Euclidean versus inertia-induced motion geometry. Two pose sequences, $q^a$ and $q^b$, are marked in both panels, and the solid curve is the shortest path between them under the metric used in that panel. (a) Euclidean trajectory space treats all directions of motion equally and ignores the articulated mass distribution. (b) The inertia matrix $M(q)$ assigns direction-dependent costs to motion, so both the shortest path between the two sequences and its length change, and the length measures the mechanical effort of transforming one sequence into the other.}
    \label{fig:flowchart}
\end{wrapfigure}

\paragraph{Limitations of Euclidean Motion Representations.}
Most motion representations operate in Euclidean configuration space, implicitly assuming an isotropic metric $\|v\|^2 = v^\top v$ in which all directions of motion are weighed equally (Figure~\ref{fig:flowchart}a). Yet articulated motion does not have this property, since moving the trunk requires substantially more mechanical effort than moving a wrist at the same kinematic speed, because the inertia loaded by trunk motion is far larger than that loaded by wrist motion. As a consequence, two trajectories that look similar in Euclidean coordinate space can correspond to very different mechanical motions (Figure~\ref{fig:flowchart}b), and trajectories that look different can correspond to the same underlying movement strategy at different body sizes. In clinical settings, the deviations associated with Parkinson's disease, including reduced stride length and altered limb swing, are interpretable as departures from the body's mechanical norms~\citep{hausdorff2009gait}, and an isotropic Euclidean distance loses the mass-distribution information that makes such deviations interpretable.

We therefore represent motion in the inertia-induced motion space, the physically realizable motions equipped with the inertia metric (Figure~\ref{fig:flowchart}b), where the distance between two motions reflects the mechanical effort needed to transform one into the other.

\subsection{Learning a Latent Embedding of the Inertia-Induced Space}

\label{sec:learning_latent}

Having defined the inertia-induced motion space (Figure~\ref{fig:flowchart}b), we seek to learn a latent representation of this space that preserves its inertia geometry.

\noindent\textbf{Latent Embedding.}
We learn an encoder
\[
z = \phi_\theta(q_{1:T}),
\]
and a decoder
\[
\hat q_{1:T} = \psi_\theta(z),
\]
where $z \in \mathbb{R}^k$ with $k \ll nT$, the dimension of a pose sequence with $n$ generalized coordinates per frame and $T$ frames.
The goal is not merely to reconstruct motion,
but to preserve the mechanical structure
defined in Section~\ref{sec:motion_space}.


\noindent\textbf{Inertia-Consistent Reconstruction.}
We measure reconstruction error under inertia geometry:
\begin{equation}
\mathcal{L}_{\text{ine}}
=
\sum_{t=1}^{T}
(\dot q_t - \dot{\hat q}_t)^\top
M(q_t)
(\dot q_t - \dot{\hat q}_t).
\end{equation}
This is the squared norm of the velocity error under the inertia metric, i.e., twice the kinetic energy of the velocity error evaluated at the ground-truth configuration. Unlike Euclidean reconstruction error, $\mathcal{L}_{\text{ine}}$ weighs each direction of motion by the body's effective inertia in that direction, so reconstruction quality is measured in the same units as the mechanical energy clinicians associate with movement. Unlike learned metrics, the inertia matrix is uniquely determined by rigid-body mechanics.




\noindent\textbf{Overall Objective.}
The final training objective is
\begin{equation}
\mathcal{L}
=
\mathcal{L}_{\text{recon}}
+
\mathcal{L}_{\text{force}}
+
\mathcal{L}_{\text{contact}}
+
\lambda_{\text{ine}}\,\mathcal{L}_{\text{ine}},
\end{equation}
where $\mathcal{L}_{\text{recon}}$, $\mathcal{L}_{\text{force}}$, and $\mathcal{L}_{\text{contact}}$ are the pose-reconstruction, force, and contact losses inherited from PhysPT~\cite{zhang2024physpt}, and $\mathcal{L}_{\text{ine}}$ is our inertia-consistent reconstruction loss replacing PhysPT's Euler-Lagrange residual term. Under this objective, the learned embedding $z$ serves as a low-dimensional representation that preserves inertia geometry.

\subsection{Scoring Function via Inertia-Induced Distance}
\label{sec:scoring}

\noindent\textbf{Scoring Overview.} Scoring takes place in the latent space. Let $\mathcal{Z}_\mathcal{N}$ be the reference embeddings of the $K$ healthy-control training subjects, and let $z^* = \phi_\theta(q^*_{1:T})$ be the embedding of a test sequence. We define the score of a test sequence as the Riemannian distance from $z^*$ to the nearest healthy reference embedding, measured under the inertia geometry of Section~\ref{sec:motion_space}. A larger score indicates a greater deviation from the healthy reference motions in terms of inertia-weighted motion differences.

The same score serves both tasks. For anomaly detection, we rank subjects by their scores and use AUROC to measure how well the ranking separates subjects with Parkinson's disease or other movement disorders from held-out healthy controls. For severity assessment, we correlate the same scores with clinical UPDRS-gait ratings on cohorts not used for training. In both cases, we compute the scores without clinical labels.

\noindent\textbf{Riemannian Distance as the Score.} Formally, the score of a test embedding $z^*$ is its distance to the nearest healthy reference embedding under the inertia geometry,
\begin{equation}
    s(z^*) = \min_{z \in \mathcal{Z}_\mathcal{N}} d_{\text{Riem}}(z^*, z),
    \label{eq:score}
\end{equation}
where $d_{\text{Riem}}$ denotes the Riemannian distance induced by the inertia metric in latent space. It is defined as the length of the shortest curve, the geodesic, connecting the two embeddings,
\begin{equation}
    d_{\text{Riem}}(z^*, z) = \min_{\gamma} \int_0^1 \big\|\dot\gamma(t)\big\|_{g(\gamma(t))}\, dt,
    \label{eq:geodesic}
\end{equation}
where $\gamma\colon[0,1]\to\mathbb{R}^k$ connects $\gamma(0)=z^*$ and $\gamma(1)=z$, $\|\dot\gamma\|_{g}=\sqrt{\dot\gamma^\top g\,\dot\gamma}$ measures each step along the curve under $g$, and $g$ is the inertia metric in latent coordinates, defined below.

\noindent\textbf{Pulling Back the Inertia Metric.}
Although we incorporate the inertia loss when learning the latent embeddings (Section~\ref{sec:learning_latent}), directly computing the L2 distance between embeddings is still a Euclidean distance rather than the Riemannian distance defined in Eq.~\ref{eq:geodesic}. The difference lies in the metric $g(z)$, which determines the length of local displacements along a path in latent space. Unlike the Euclidean metric, which weights all latent directions equally, $g(z)$ accounts for the inertia associated with changes in the decoded poses. Figure~\ref{fig:pipeline} (right) illustrates the resulting difference between Euclidean and Riemannian distances. To obtain $g(z)$, we pull back the inertia metric from pose space to latent space through the decoder~\cite{arvanitidis2017latent,arvanitidis2020geometrically}.

Specifically, a latent displacement $\Delta z$ at an embedding $z$ moves the decoded poses $\hat q=\psi_\theta(z)$ by $\Delta q = J(z)\Delta z$, where $J(z)$ is the Jacobian of the decoder at $z$, computed by forward-mode automatic differentiation. Using the same inertia weighting as in $\mathcal{L}_{\text{ine}}$ (Section~\ref{sec:learning_latent}), the squared displacement $\Delta q^\top M(\hat q)\Delta q$ can be expressed in latent coordinates as $\Delta z^\top g(z)\Delta z$, where
\begin{equation}
    g(z) = J(z)^\top M(\hat q)\, J(z),
    \label{eq:pullback}
\end{equation}
and $g(z)$ is the inertia metric in latent coordinates at $z$. It is a $k\times k$ symmetric positive-definite matrix that varies with $z$ because both $J$ and $M(\hat q)$ depend on $z$. Here $M(\hat q)$ is block-diagonal, with one block $M(\hat q_t)$ for each of the $T$ decoded frames. The quadratic form $\Delta q^\top M(\hat q)\Delta q$ therefore sums the per-frame terms, as in $\mathcal{L}_{\text{ine}}$. The matrix $g(z)$ is $k\times k$ and symmetric. It is positive definite if $J(z)$ has full column rank, which we assume for the trained decoder~\cite{arvanitidis2020geometrically}. It varies with $z$ because both $J$ and $M(\hat q)$ depend on $z$. This computation is exact and is performed only once for each embedding.

\noindent\textbf{Graph-Based Approximation of Riemannian Distance.}
Computing the Riemannian distance in Eq.~\ref{eq:geodesic} requires optimizing over continuous curves, which is computationally prohibitive in practice. We therefore approximate the Riemannian distance in Eq.~\ref{eq:geodesic} using a shortest path on a graph~\cite{tenenbaum2000global,chen2019fast}. 

The graph is a $\kappa$-nearest-neighbor graph ($\kappa{=}20$) with the reference embeddings in $\mathcal{Z}_\mathcal{N}$ and the embeddings of one test subject as nodes. The metric $g$ is computed exactly at every node using Eq.~\ref{eq:pullback}. Each edge is a straight segment between two nodes $z_i$ and $z_j$. Its length under $g$ is given by the integral in Eq.~\ref{eq:geodesic} along the segment. We approximate this length using the average of the endpoint metrics, yielding $\sqrt{\Delta z^\top \tfrac{1}{2}\big(g(z_i)+g(z_j)\big)\Delta z}$ where $\Delta z = z_i - z_j$. The sum of edge lengths along the shortest path between two nodes approximates their Riemannian distance. Running Dijkstra's algorithm with all reference embeddings as sources returns the minimum in Eq.~\ref{eq:score} for every node in a single pass. We build the graph separately for each test subject so that a subject's score does not depend on other test subjects. All reported REINS results use this procedure. Appendix~\ref{sec:graph_metric} describes a constant approximation of $g$ fitted once on the reference embeddings and reports sensitivity to the graph construction.

\section{Experiments}
\label{sec:experiments}


\noindent\textbf{Datasets and Tasks.} 
We use CARE-PD~\cite{adeli2025care}, a multi-cohort Parkinson's-disease gait benchmark. Its DNE (Digitized Neurological Examination) cohort contains 28 healthy controls (HC), 30 Parkinson's-disease (PD) subjects, and 39 subjects with other movement-related diseases, with no UPDRS-gait severity labels. Four other cohorts, PD-GaM, BMCLab, T-SDU-PD, and 3DGait, have subject-level UPDRS-gait ratings and were collected at different sites under different protocols. We fine-tune the PhysPT encoder-decoder described in Section~\ref{sec:learning_latent}~\cite{zhang2024physpt} using only DNE healthy controls. We evaluate HC vs.\ PD discrimination on DNE and, without further fine-tuning, correlate the scores with UPDRS-gait ratings on the four labeled cohorts.

\noindent\textbf{Baselines.}
Employing the same PhysPT-pretrained encoder-decoder backbone~\cite{zhang2024physpt}, we compare three variations of our approach that differ in the geometry of the resulting latent space and the scoring distance used at evaluation (Figure~\ref{fig:pipeline}):

\begin{itemize}[leftmargin=1.5em, itemsep=2pt, topsep=2pt]
    \item \textit{Data-driven Euclidean} fine-tunes with $\mathcal{L}_{\text{recon}}$ only when obtaining the latent representation and scores each window by its Mahalanobis distance to the mean of the reference windows, with covariance estimated from the same windows using Ledoit-Wolf shrinkage~\cite{ledoit2004well}. 
    \item \textit{Physics-supervised Euclidean} fine-tunes with the original PhysPT objective ($\mathcal{L}_{\text{recon}}+\mathcal{L}_{\text{force}}+\mathcal{L}_{\text{contact}}+\mathcal{L}_{\text{euler}}$), with the same Euclidean read-out.
    \item \textit{REINS (ours)} fine-tunes with the REINS objective ($\mathcal{L}_{\text{recon}}+\mathcal{L}_{\text{force}}+\mathcal{L}_{\text{contact}}+\lambda_{\text{ine}}\mathcal{L}_{\text{ine}}$, $\lambda_{\text{ine}}=0.1$), in which our inertia-consistent loss $\mathcal{L}_{\text{ine}}$ replaces the Euler-Lagrange residual. Our REINS defines an inertia-induced Riemannian latent that is scored by Riemannian distance under the pulled-back inertia metric of Section~\ref{sec:scoring}.
\end{itemize}

We additionally consider three published motion encoders, VPoser~\cite{pavlakos2019expressive}, GaitForeMer~\cite{endo2022gaitforemer}, and MotionBERT~\citep{zhu2023motionbert}, with frozen backbones, scored as described in Appendix~\ref{sec:baselines}. 

\noindent\textbf{Implementation Details.}
All conditions are fine-tuned from the PhysPT checkpoint pretrained on AMASS~\cite{zhang2024physpt,mahmood2019amass}, using Adam with learning rate $10^{-6}$, weight decay $10^{-4}$, batch size 16, and a step learning-rate schedule (decay factor 0.8 every 5 epochs). All conditions are fine-tuned for up to 50 epochs, with early stopping on held-out HC mean per-joint position error (MPJPE) (patience 10, minimum 20 epochs). All optimization hyperparameters are held fixed across conditions and dataset-size regimes so that downstream differences are attributable to the loss configuration or training-set size rather than tuning. Experiments use a single NVIDIA RTX 3090 GPU. Every recording is split into 16-frame windows at 30 fps~\cite{zhang2024physpt}. 

For input, each window is one input sequence $q_{1:T}$ with $T{=}16$ for both fine-tuning and evaluation. At evaluation, every window of a subject is embedded and scored as in Section~\ref{sec:scoring}, and the subject's score is the median of its window-level scores. For latent space, before scoring we reduce the latent dimensionality using PCA fitted on the reference embeddings $\mathcal{Z}_\mathcal{N}$, retaining 95\% of the variance, and remove the leading component, which corresponds to global translation along the walking direction.

\noindent\textbf{Evaluation Protocol.}
Of the 28 DNE healthy controls, 8 are held out for testing, and $K$ of the remaining 20 are used for fine-tuning and to form the reference embeddings $\mathcal{Z}_\mathcal{N}$ defined in Section~\ref{sec:scoring}. PD and other-disease subjects are used only as anomalies at evaluation. Subject-level AUROC is computed on the 8 held-out HC against the 30 PD subjects (HC/PD) and against the 30 PD plus 39 other-disease subjects (HC/all). The main results use $K{=}20$, and Section~\ref{sec:riem_euc} reduces $K$ to study data scarcity. We use one split of the healthy controls for the main results (the main split). Table~\ref{tab:main} reports 95\% bootstrap confidence intervals over the test subjects, with the model held fixed (Appendix~\ref{sec:paired}). In Section~\ref{sec:riem_euc}, we vary the training subjects across random splits. Appendix~\ref{sec:paired} describes the bootstrap and random-split procedures and reports confidence intervals, HC/all results, and the factorial ablation for all $K$, together with a paired bootstrap test of the differences between methods. On the four UPDRS-labeled cohorts we report Spearman $\rho$ and Kendall $\tau$-b between subject-level scores and UPDRS-gait ratings to assess whether the score increases with clinical severity.

\noindent\textbf{Experiments Overview.} In Section~\ref{sec:riem_euc}, we test whether the Riemannian score separates PD from HC better than the Euclidean readouts and three published motion encoders (Table~\ref{tab:main}), disentangle the contributions of loss and scoring (Table~\ref{tab:factorial}), and evaluate performance under limited healthy training data (Table~\ref{tab:scarcity}). In Section~\ref{sec:zeroshot}, we correlate the resulting scores with UPDRS-gait ratings for patients from four cohorts in a zero-shot setting (Table~\ref{tab:zeroshot}).

\subsection{Pathological Motion under Riemannian vs Euclidean Latent Geometry}

\label{sec:riem_euc}

\textbf{\textit{TL;DR:}} REINS changes two things relative to a standard latent space: the encoder is fine-tuned with the inertia-consistent loss, and subjects are scored by Riemannian rather than Euclidean distance. We test whether either change alone separates subjects with Parkinson's disease from healthy controls on DNE. The REINS encoder with Euclidean scoring reaches 0.62 AUROC and the standard PhysPT encoder with Riemannian scoring reaches 0.59, while combining the two changes gives 0.91 (Tables~\ref{tab:main} and~\ref{tab:factorial}). For Riemannian scoring, using the decoder Jacobian alone gives 0.73 AUROC, which rises to 0.91 when the inertia matrix is included. The separation holds even with a single healthy training subject (Table~\ref{tab:scarcity}). Detailed experimental results and analyses are presented below.

\noindent\textbf{Comparison with Baselines.} Table~\ref{tab:main} (left) shows that on the main split REINS separates PD from HC, with AUROC 0.9083 on HC/PD and 0.8768 on HC/all, while the two Euclidean readouts of the same backbone remain near chance. The physics-supervised readout is near chance despite being trained with the full PhysPT physics objective, suggesting that trajectory-level physics losses alone do not make pathological motion separable in the latent space. On the main split, REINS improves AUROC over MotionBERT by 0.14 on HC/PD and 0.12 on HC/all.

\noindent\textbf{Sensitivity to the Subject Split.} The main split uses 20 healthy controls for training and holds out 8 for testing. To assess sensitivity to this split, we repeat the $K{=}20$ experiment on ten random splits and fine-tune the model on each split. The right half of Table~\ref{tab:main} reports the mean and standard deviation of AUROC across the ten splits. Across the tested splits, REINS achieves mean AUROCs of 0.8804 on HC/PD and 0.8663 on HC/all, with standard deviations of 0.027 and 0.020, respectively, compared with mean AUROCs of 0.7621 and 0.7451 for MotionBERT.

\noindent\textbf{Separating the Loss from the Scoring.} Table~\ref{tab:main} changes the fine-tuning objective and the scoring distance together, so it is unclear how much each contributes to the gain. In Table~\ref{tab:factorial}, we evaluate each encoder with three scoring distances. The two encoders are the physics-supervised encoder, trained with the original PhysPT objective, and the REINS encoder, trained with $\mathcal{L}_{\text{ine}}$. The highest AUROC requires both the inertia loss for training and the inertia metric for scoring. Scoring the physics-supervised encoder with the Riemannian distance raises AUROC from 0.5167 to 0.5917. The REINS encoder with the Euclidean readout gives 0.6208, while combining the REINS encoder and Riemannian distance gives 0.9083. For the REINS encoder, the $J^\top J$ and $J^\top M J$ Riemannian distances yield AUROCs of 0.7333 and 0.9083. Adding the inertia matrix therefore improves AUROC by 0.18, compared with 0.11 from replacing the Euclidean readout with the $J^\top J$ Riemannian distance. This is consistent with the clinical interpretation of bradykinesia and reduced stride length as departures from the body's mechanical norms~\cite{hausdorff2009gait}, which the inertia geometry encodes through direction-dependent motion costs. Appendix~\ref{sec:ablation} compares pose and kinetic-energy errors to assess how the fine-tuning objective affects the decoded motion. The additional conditions include training from scratch, using the pretrained checkpoint without fine-tuning, and fine-tuning with the Euler--Lagrange residual removed from the PhysPT objective. Replacing the residual with $\mathcal{L}_{\text{ine}}$ reduces kinetic-energy error from 193.4\% to 22.7\% and lowers MPJPE by 0.19~mm.

\begin{table*}[t]
\centering
\small
\caption{Anomaly detection AUROC with $K{=}20$ HC training subjects. Left: the main split, with 95\% bootstrap confidence intervals over the evaluation subjects. Right: mean $\pm$ standard deviation over ten random splits of the healthy controls into training and held-out subjects. HC/PD = HC vs PD, HC/all = HC vs PD$+$other. Riemannian scoring in our $M$-induced Riemannian latent space outperforms Euclidean baselines and three published motion encoders.}
\label{tab:main}
\setlength{\tabcolsep}{4pt}
\renewcommand{\arraystretch}{1.1}
\begin{tabular}{lcccc}
\toprule
 & \multicolumn{2}{c}{\textbf{Main split} [95\% CI]} & \multicolumn{2}{c}{\textbf{Ten random splits} (mean $\pm$ sd)} \\
\cmidrule(lr){2-3}\cmidrule(lr){4-5}
\textbf{Method} & HC/PD & HC/all & HC/PD & HC/all \\
\midrule
\multicolumn{5}{l}{\emph{Euclidean latent space}\,   \,distance: Euclidean} \\
\quad Data-driven        & 0.5750 [0.3792, 0.7625] & 0.6087 [0.4312, 0.7736] & 0.5633 $\pm$ 0.050 & 0.5988 $\pm$ 0.034 \\
\quad Physics-supervised & 0.5167 [0.3167, 0.7125] & 0.5380 [0.3587, 0.7156] & 0.5242 $\pm$ 0.049 & 0.5442 $\pm$ 0.033 \\
\midrule
\multicolumn{5}{l}{\emph{Published motion encoders}\,   \,distance: Riemannian} \\
\quad VPoser~\cite{pavlakos2019expressive}   & 0.5708 [0.3750, 0.7708] & 0.6051 [0.4312, 0.7790] & 0.5888 $\pm$ 0.046 & 0.5998 $\pm$ 0.031 \\
\quad GaitForeMer~\cite{endo2022gaitforemer} & 0.7125 [0.5250, 0.8792] & 0.7337 [0.5815, 0.8768] & 0.7071 $\pm$ 0.033 & 0.7205 $\pm$ 0.036 \\
\quad MotionBERT~\cite{zhu2023motionbert} & 0.7667 [0.5458, 0.9208] & 0.7572 [0.5870, 0.8986] & 0.7621 $\pm$ 0.025 & 0.7451 $\pm$ 0.029 \\
\midrule
\multicolumn{5}{l}{\emph{Riemannian latent space} ($M$-induced, ours)\,   \,distance: Riemannian} \\
\quad REINS              & \textbf{0.9083} [0.8042, 0.9833] & \textbf{0.8768} [0.7736, 0.9529] & \textbf{0.8804} $\pm$ 0.027 & \textbf{0.8663} $\pm$ 0.020 \\
\bottomrule
\end{tabular}
\end{table*}

\begin{table}[t]
\centering
\caption{Factorial ablation of the fine-tuning objective and the scoring distance. Subject-level HC/PD AUROC at $K{=}20$ with 95\% bootstrap intervals. The rows correspond to the physics-supervised encoder (without $\mathcal{L}_{\text{ine}}$) and the REINS encoder (with $\mathcal{L}_{\text{ine}}$) in Table~\ref{tab:main}. The $J^\top J$ Riemannian distance uses the same procedure without the inertia matrix.}
\label{tab:factorial}
\setlength{\tabcolsep}{4pt}
\renewcommand{\arraystretch}{1.3}
\small
\begin{tabular}{lccc}
\toprule
\textbf{Fine-tuning} & \textbf{Euclidean} & \textbf{$J^\top J$ Riemannian} & \textbf{$J^\top M J$ Riemannian} \\
\midrule
Physics-supervised (w/o $\mathcal{L}_{\text{ine}}$) & 0.5167 [0.3167, 0.7125] & 0.5542 [0.3583, 0.7750] & 0.5917 [0.3833, 0.7708] \\
REINS (w/ $\mathcal{L}_{\text{ine}}$)    & 0.6208 [0.4250, 0.8000] & 0.7333 [0.5375, 0.8667] & \textbf{0.9083} [0.8042, 0.9833] \\
\bottomrule
\end{tabular}
\end{table}

\noindent\textbf{Physical Prior under Clinical Data Scarcity.} A defining property of clinical motion data is that it is fundamentally limited in scale: enrolling additional subjects requires clinical recruitment, and per-subject recording is constrained by patient comfort and protocol cost. The relevant question for clinical deployment is therefore not how a method performs at the size of a curated mocap corpus, but how it degrades as the available training data shrinks. We evaluate this by re-training each method on progressively reduced fractions of the DNE training cohort and re-measuring downstream HC vs.\ PD discrimination. The inertia matrix $M(q)$ is mechanically determined and is not estimated from the training set. We hypothesize that it remains a useful inductive bias as the number of healthy training subjects decreases, when purely data-driven fine-tuning has less data from which to learn structure.

Following the experimental protocol described above, we draw $K$ training subjects from the 20 non-held-out HC, and their windows form the reference embeddings $\mathcal{Z}_\mathcal{N}$. For each $K$, the model is fine-tuned from the same PhysPT pretrained checkpoint, with all training and optimization hyperparameters held fixed across $K$ so that any difference in downstream performance is attributable to training-set size rather than tuning. We use $K \in \{1, 3, 5, 10, 15, 20\}$, where the training subjects are nested ($K{=}1 \subset K{=}3 \subset K{=}5 \subset K{=}10 \subset K{=}15 \subset K{=}20$) so that smaller-$K$ runs see a strict subset of the data seen by larger-$K$ runs and cross-$K$ comparisons are not confounded by sampling variance. {Table~\ref{tab:scarcity} reports HC/PD AUROC for each $K$, with $K{=}20$ the main setting of Table~\ref{tab:main}.}

Across all training-set sizes, REINS yields higher AUROC than both Euclidean baselines (Table~\ref{tab:scarcity}). The gains range from 0.22 to 0.33 over the data-driven baseline and from 0.28 to 0.39 over the physics-supervised baseline, with no monotonic trend in $K$. Four Euclidean results fall below 0.5 (0.43--0.48). For 8 HC and 30 PD test subjects, the approximate 95\% null range for AUROC is $[0.28, 0.72]$. These results are therefore consistent with chance, and their confidence intervals include 0.5 (Table~\ref{tab:scarcity_ci}). REINS, by contrast, retains AUROC 0.7375 at $K{=}1$. Since the inertia matrix $M(q)$ is determined by rigid-body mechanics rather than estimated from data, a representation grounded in this geometry continues to organize motion meaningfully even when the optimizer sees almost no healthy examples. At $K{=}1$, the reference embeddings $\mathcal{Z}_\mathcal{N}$ contain 97 windows from a single subject selected using the split seed, so the reported result depends on the choice of reference subject. In a clinical-deployment regime, where each additional healthy control subject costs recruitment and protocol time, this property is what makes the representation usable.

\begin{table}[t]
\centering
\caption{Data efficiency: HC/PD AUROC across HC training-subject count $K$.
Even at $K{=}1$, our REINS ($M$-induced Riemannian) substantially outperforms both Euclidean baselines. $K{=}20$ is the main reported setting (Table~\ref{tab:main}). Table~\ref{tab:scarcity_ci} reports the confidence intervals and HC/all results.}
\label{tab:scarcity}
\setlength{\tabcolsep}{6pt}
\renewcommand{\arraystretch}{1}
\begin{tabular}{lcccccc}
\toprule
\textbf{Method} & $K{=}20$ & $K{=}15$ & $K{=}10$ & $K{=}5$ & $K{=}3$ & $K{=}1$ \\
\midrule
\multicolumn{7}{l}{\emph{Euclidean latent space}\,   \,distance: Euclidean} \\
\quad Data-driven           & 0.5750 & 0.5708 & 0.5333 & 0.5250 & 0.5375 & 0.4292 \\
\quad Physics-supervised    & 0.5167 & 0.5250 & 0.5125 & 0.4583 & 0.4792 & 0.4417 \\
\midrule
\multicolumn{7}{l}{\emph{Riemannian latent space} ($M$-induced, ours)\,   \,distance: Riemannian} \\
\quad REINS         & \textbf{0.9083} & \textbf{0.8458} & \textbf{0.8125} & \textbf{0.7792} & \textbf{0.7583} & \textbf{0.7375} \\
\bottomrule
\end{tabular}
\end{table}

\subsection{Zero-Shot Severity Correlation across Heterogeneous Cohorts}
\label{sec:zeroshot}

A clinically deployed gait phenotyping system is evaluated on cohorts and acquisition sites unseen at training time. We test whether REINS, fine-tuned only on DNE healthy controls, produces severity scores that correlate with UPDRS-gait ratings on four external UPDRS-gait-labeled cohorts, PD-GaM, BMCLab, T-SDU-PD, and 3DGait~\cite{adeli2025care}, without any fine-tuning, label exposure, or per-cohort calibration on these target cohorts. The hypothesis is that a representation grounded in body-level inertia, which is shared across cohorts and sites, transfers more reliably than representations whose structure is learned from cohort-specific kinematic patterns. REINS scores correlate positively with UPDRS-gait ratings on all four cohorts, with significant Spearman correlations on three. Neither Euclidean readout shows a significant correlation on any cohort (Table~\ref{tab:zeroshot}).

\noindent\textbf{Zero-Shot Protocol.} For each test subject in the four target cohorts, we extract REINS embeddings using the same DNE-fine-tuned encoder as in Section~\ref{sec:riem_euc}, and take the median of the window-level Riemannian distances to the DNE HC reference embeddings $\mathcal{Z}_\mathcal{N}$ as the subject-level score. Each target subject's graph contains only the reference windows and that subject's windows, as in Section~\ref{sec:scoring}. We then measure the rank correlation between subject-level scores and ground-truth UPDRS-gait ratings using Spearman $\rho$ and Kendall $\tau$-b. We report the latter because UPDRS-gait is ordinal and contains ties at each integer level. No supervision from the four cohorts enters the encoder or the score.  Asterisks in Table~\ref{tab:zeroshot} indicate rank correlations significantly different from zero ($p{<}0.05$, Appendix~\ref{sec:paired}).

\noindent\textbf{Results.} Table~\ref{tab:zeroshot} shows that the Riemannian score correlates with UPDRS-gait severity on cohorts not used for training. REINS produces positive rank correlations on all four cohorts, while both Euclidean baselines, despite sharing the same encoder backbone, hover in a substantially lower range. For the Riemannian score, Spearman correlation is significant ($p{<}0.05$) on PD-GaM, BMCLab and 3DGait, but not on T-SDU-PD ($n{=}14$). Kendall $\tau$-b is significant on PD-GaM and BMCLab, whereas no Euclidean baseline shows a significant correlation on any cohort. The fact that this gap holds on every target cohort indicates that the underlying severity signal carried by the Riemannian distance is not specific to the cohort the encoder was trained on. The correlations also do not collapse despite substantial heterogeneity across PD-GaM, BMCLab, T-SDU-PD, and 3DGait in subject count, recording protocol, and severity distribution. This is the behaviour one would expect if the score reflects departures from the body's mechanical norms~\cite{hausdorff2009gait} rather than cohort-specific kinematic patterns, since rigid-body dynamics depend on the articulated body itself rather than on the camera or laboratory in which it is recorded.

\begin{table}[t]
\centering
\caption{Zero-shot UPDRS-gait severity correlation on four heterogeneous UPDRS-gait-labeled cohorts.
The model is fine-tuned only on DNE healthy controls ($K{=}20$) with no labels or samples from the target cohorts entering the encoder. Subject-level Spearman $\rho$ and Kendall $\tau$-b between Riemannian distance to the DNE HC reference and ground-truth UPDRS-gait scores. Asterisks indicate rank correlations significantly different from zero ($p{<}0.05$, Appendix~\ref{sec:paired}). The subject is the unit of analysis, and $n$ denotes the number of subjects in each cohort.}
\label{tab:zeroshot}
\setlength{\tabcolsep}{4pt}
\renewcommand{\arraystretch}{1.3}
\small
\begin{tabular}{lcccccccc}
\toprule
\textbf{Method} & \multicolumn{2}{c}{\textbf{PD-GaM} ($n{=}30$)}
                & \multicolumn{2}{c}{\textbf{BMCLab} ($n{=}23$)}
                & \multicolumn{2}{c}{\textbf{T-SDU-PD} ($n{=}14$)}
                & \multicolumn{2}{c}{\textbf{3DGait} ($n{=}43$)} \\
\cmidrule(lr){2-3}\cmidrule(lr){4-5}\cmidrule(lr){6-7}\cmidrule(lr){8-9}
& $\rho$ & $\tau$ & $\rho$ & $\tau$ & $\rho$ & $\tau$ & $\rho$ & $\tau$ \\
\midrule
\multicolumn{9}{l}{\emph{Euclidean latent space}\,   \,distance: Euclidean} \\
\quad Data-driven         & 0.2802 & 0.1648 & 0.2200 & 0.1972
                          & 0.2569 & 0.2405 & 0.2138 & 0.1891 \\
\quad Physics-supervised  & 0.2510 & 0.1405 & 0.2895 & 0.2648
                          & 0.2337 & 0.1243 & 0.1700 & 0.1567 \\
\midrule
\multicolumn{9}{l}{\emph{Riemannian latent space} ($M$-induced, ours)\,   \,distance: Riemannian} \\
\quad REINS               & \textbf{0.5649*} & \textbf{0.4582*}
                          & \textbf{0.5742*} & \textbf{0.4400*}
                          & \textbf{0.4308} & \textbf{0.3078}
                          & \textbf{0.3154*} & \textbf{0.2132} \\
\bottomrule
\end{tabular}
\end{table}

\section{Conclusion}
We present REINS, a motion representation that preserves the biomechanical structure of motion in the learned latent space by using the Riemannian geometry induced by articulated rigid-body inertia. On the CARE-PD~\cite{adeli2025care} benchmark, REINS substantially outperforms data-driven and physics-supervised Euclidean baselines in distinguishing healthy controls from subjects with Parkinson's disease. To test whether the advantage holds when healthy training data are scarce, we reduce the number of healthy training subjects from 20 to one. REINS still outperforms the Euclidean baselines with a single training subject. Its scores also correlate with UPDRS-gait ratings in zero-shot evaluations on four heterogeneous external cohorts without per-cohort calibration. These results therefore support using the inertia metric as a common basis for comparing gait across training-set sizes and acquisition sites. We compute the inertia matrix from the body model and pose using rigid-body mechanics, independently of the training data. REINS uses this geometry to characterize how a person's gait departs from healthy movement, a comparison that is central to clinical gait assessment.

\noindent\textbf{Limitations and Future Directions.}
We compute the inertia matrix using the mean SMPL shape, as CARE-PD does not provide subject-specific shape parameters~\cite{adeli2025care}. Finer-grained personalization of mass distribution from clinically measurable anthropometrics (limb-length, body composition) may further improve the calibration of motion costs at the subject level. Our scoring is computed against a healthy-control reference cohort drawn from one dataset (CARE-PD/DNE), and broader reference cohorts spanning age, body size, and recording protocol would further strengthen the population-level calibration of the resulting severity score. Further work could quantify how uncertainty in monocular pose estimates~\cite{shin2024wham} affects $M(q)$ and the resulting scores. Our evaluation uses one benchmark and one fine-tuned backbone~\cite{zhang2024physpt}, and the protocol can be extended to other motion priors. Beyond the inertia matrix studied here, muscle moment arms $R(q)=\partial l_{MT}/\partial q$, computed from pose in a scaled musculoskeletal model~\cite{sherman2013moment}, could be used to construct an additional mechanically meaningful quadratic form. Finally, the representation is currently evaluated through subject-level summary scores aggregated as the median of window-level Riemannian distances. Per-window or temporally resolved variants of the same Riemannian distance may surface stride-level patterns that carry clinical information but are averaged out at the subject level.

\begin{ack}
This work was supported by Michigan State University.
\end{ack}

\bibliographystyle{plainnat}
\bibliography{references}


\appendix

\section{Ablation Study}
\label{sec:ablation}

We ablate the contribution of the inertia-consistent reconstruction loss $\mathcal{L}_{\text{ine}}$ by comparing six conditions that vary which losses are active during fine-tuning. All conditions share the same encoder--decoder architecture; conditions C, D, and F correspond to the three setups defined in Section~\ref{sec:experiments}, while A, B, and E provide additional reference points.
\begin{itemize}[leftmargin=1.5em, itemsep=2pt, topsep=2pt]
    \item \textit{A: From scratch.} No pretraining; trained on DNE with $\mathcal{L}_{\text{recon}}+\mathcal{L}_{\text{force}}+\mathcal{L}_{\text{contact}}+\mathcal{L}_{\text{euler}}$.
    \item \textit{B: Pretrained, no fine-tuning.} Frozen PhysPT pretrained checkpoint, evaluated directly on DNE.
    \item \textit{C: Pretrained + $\mathcal{L}_{\text{recon}}$ only.} Equivalent to \textit{Data-driven Euclidean}.
    \item \textit{D: Pretrained + PhysPT objective.} $\mathcal{L}_{\text{recon}}+\mathcal{L}_{\text{force}}+\mathcal{L}_{\text{contact}}+\mathcal{L}_{\text{euler}}$. Equivalent to \textit{Physics-supervised Euclidean}.
    \item \textit{E: Pretrained + PhysPT objective without $\mathcal{L}_{\text{euler}}$.} $\mathcal{L}_{\text{recon}}+\mathcal{L}_{\text{force}}+\mathcal{L}_{\text{contact}}$, isolating the contribution of the Euler--Lagrange residual loss.
    \item \textit{F: Pretrained + REINS objective (ours).} $\mathcal{L}_{\text{recon}}+\mathcal{L}_{\text{force}}+\mathcal{L}_{\text{contact}}+\lambda_{\text{ine}}\mathcal{L}_{\text{ine}}$. Equivalent to \textit{REINS (ours)}.
\end{itemize}
For each condition we report physical reconstruction quality (MPJPE, ACCEL, Euler-Lagrange residual, kinetic-energy (KE) error) on the DNE held-out test set. The relative KE error is $|\bar E_{\text{pred}}-\bar E_{\text{GT}}|/\bar E_{\text{GT}}$, where $\bar E_{\text{pred}}$ and $\bar E_{\text{GT}}$ are the mean predicted and ground-truth kinetic energies over the test set, respectively.

In Table~\ref{tab:ablation}, there are three patterns that need to be discussed. First, comparing condition C (recon-only, 82.1\% KE relative error) with condition F (REINS, 22.7\% KE relative error) shows that adding $\mathcal{L}_{\text{ine}}$ reduces the predicted-vs.-GT kinetic-energy gap by roughly a factor of four, with a 0.74 mm increase in MPJPE (9.05 vs.\ 9.79 mm). Second, the original PhysPT objective (condition D) is the worst on KE (193.4\%) despite achieving the lowest Euler-Lagrange residual (32.41), indicating that minimizing the Euler-Lagrange residual at the trajectory level does not by itself produce decoded motion whose kinetic energy matches the body's. Third, removing $\mathcal{L}_{\text{euler}}$ (condition E, 101.3\% KE relative error) recovers part of the gap, and replacing it with our inertia-consistent loss (condition F, 22.7\% KE relative error) closes most of the remainder. This phenomenon suggests that enforcing physics as a trajectory-level penalty leaves the latent geometry unconstrained, while supervising reconstruction under the inertia-weighted norm aligns the decoder with the actual mass distribution of an articulated body, recovering kinetic-energy fidelity at the comparable pose-reconstruction quality. The effect on the latent geometry is measured directly by Table~\ref{tab:factorial}.

\begin{table}[!htbp]
\centering
\caption{Ablation on training conditions A-F. Conditions defined in Section~\ref{sec:ablation}.}
\label{tab:ablation}
\small
\begin{tabular}{@{}c ccccc ccc@{}}
\toprule
& \multirow{2}{*}{\shortstack{Val \\ MPJPE \\ (mm)}}
  & \multirow{2}{*}{\shortstack{Test \\ MPJPE \\ (mm)}}
  & \multirow{2}{*}{\shortstack{PA- \\ MPJPE \\ (mm)}}
  & \multirow{2}{*}{\shortstack{ACCEL \\ (mm/s$^2$)}}
  & \multirow{2}{*}{\shortstack{EL \\ Residual}}
  & \multicolumn{3}{c}{Kinetic Energy (J)} \\
  \cmidrule(l){7-9}
Exp & & & & & & Pred & GT & rel error \\
\midrule
A & 30.8 & 32.44 & 28.67 & 3.19 & 36.51 & 16.76 & 9.94 & 68.6\% \\
B & --   & 31.60 & 21.66 & 3.21 & 39.52 & 44.98 & 9.94 & 352.5\% \\
C & 8.4  & 9.05  & 7.07  & 2.50 & 38.42 & 18.10 & 9.94 & 82.1\% \\
D & 9.5  & 9.98  & 7.91  & 2.63 & 32.41 & 29.17 & 9.94 & 193.4\% \\
E & 9.4  & 9.81  & 7.67  & 2.54 & 38.31 & 20.02 & 9.94 & 101.3\% \\
F & 9.4  & 9.79 & 7.73 & 2.53 & 37.10 & \textbf{7.68} & 9.94 & \textbf{22.7\%} \\
\bottomrule
\end{tabular}
\end{table}

\section{Statistical Protocol and Additional Results}
\label{sec:paired}

This appendix gives the bootstrap and split procedures behind the intervals and the mean $\pm$ sd in the main tables, and reports results for all $K$. The main split is the default split in the released code. To compute the percentile bootstrap confidence intervals, we keep the model fixed, draw subjects with replacement separately from the held-out HC and the positive group, and recompute AUROC. We repeat this procedure 2000 times. The $K{=}20$ setting is repeated over ten random splits of the healthy controls, with the model fine-tuned anew on each split and reported as mean $\pm$ standard deviation. Rank correlations in Table~\ref{tab:zeroshot} are tested against the null of zero correlation using the $p$-values returned by SciPy's \texttt{spearmanr} and \texttt{kendalltau} ($\tau$-b), with the subject as the unit. Table~\ref{tab:scarcity_ci} reports the confidence intervals and HC/all results corresponding to Table~\ref{tab:scarcity}. Table~\ref{tab:factorial_k} reports the factorial ablation for each $K$. Across all $K$, AUROC is highest with the $J^\top M J$ Riemannian distance, followed by the $J^\top J$ Riemannian distance and the Euclidean readout. The encoder trained without $\mathcal{L}_{\text{ine}}$ performs near chance with all three distances. Table~\ref{tab:paired} reports paired bootstrap differences between REINS and each Euclidean baseline, using the same drawn subjects for both methods in each comparison. All intervals exclude zero.

\begin{table}[!htbp]
\centering
\caption{Data efficiency with 95\% bootstrap intervals: HC/PD and HC/all AUROC for different numbers of HC training subjects $K$.}
\label{tab:scarcity_ci}
\footnotesize
\setlength{\tabcolsep}{3pt}
\newcommand{\ci}[2]{\begin{tabular}[c]{@{}c@{}}#1\\[-1pt]{\scriptsize[#2]}\end{tabular}}
\begin{tabular}{@{}lcccccc@{}}
\toprule
\textbf{Method} & $K{=}20$ & $K{=}15$ & $K{=}10$ & $K{=}5$ & $K{=}3$ & $K{=}1$ \\
\midrule
\multicolumn{7}{@{}l}{\textit{HC/PD}} \\
Data-driven        & \ci{0.5750}{0.3792, 0.7625} & \ci{0.5708}{0.3500, 0.7750} & \ci{0.5333}{0.3208, 0.7458} & \ci{0.5250}{0.3083, 0.7292} & \ci{0.5375}{0.3417, 0.7417} & \ci{0.4292}{0.2292, 0.6417} \\
Physics-supervised & \ci{0.5167}{0.3167, 0.7125} & \ci{0.5250}{0.3250, 0.7250} & \ci{0.5125}{0.3167, 0.7250} & \ci{0.4583}{0.2750, 0.6500} & \ci{0.4792}{0.2708, 0.6500} & \ci{0.4417}{0.2542, 0.6458} \\
REINS              & \ci{\textbf{0.9083}}{0.8042, 0.9833} & \ci{\textbf{0.8458}}{0.6917, 0.9458} & \ci{\textbf{0.8125}}{0.6208, 0.9250} & \ci{\textbf{0.7792}}{0.6208, 0.9792} & \ci{\textbf{0.7583}}{0.6125, 0.9000} & \ci{\textbf{0.7375}}{0.5125, 0.9292} \\
\midrule
\multicolumn{7}{@{}l}{\textit{HC/all}} \\
Data-driven        & \ci{0.6087}{0.4312, 0.7736} & \ci{0.5707}{0.3986, 0.7736} & \ci{0.5344}{0.3261, 0.7283} & \ci{0.5163}{0.3351, 0.7246} & \ci{0.5217}{0.3496, 0.7101} & \ci{0.4149}{0.2283, 0.6812} \\
Physics-supervised & \ci{0.5380}{0.3587, 0.7156} & \ci{0.5145}{0.3352, 0.7029} & \ci{0.5036}{0.3315, 0.6938} & \ci{0.4620}{0.3062, 0.6341} & \ci{0.4511}{0.2953, 0.6087} & \ci{0.4149}{0.2428, 0.6214} \\
REINS              & \ci{\textbf{0.8768}}{0.7736, 0.9529} & \ci{\textbf{0.8351}}{0.6920, 0.9330} & \ci{\textbf{0.7917}}{0.6522, 0.8768} & \ci{\textbf{0.7428}}{0.6250, 0.8406} & \ci{\textbf{0.7264}}{0.6087, 0.8279} & \ci{\textbf{0.7011}}{0.5833, 0.8225} \\
\bottomrule
\end{tabular}
\end{table}

\begin{table*}[!htbp]
\centering
\caption{Factorial ablation across the number of HC training subjects $K$. Subject-level HC/PD AUROC with 95\% bootstrap intervals.}
\label{tab:factorial_k}
\scriptsize
\setlength{\tabcolsep}{3pt}
\newcommand{\ci}[2]{\begin{tabular}[c]{@{}c@{}}#1\\[-1pt]{\tiny[#2]}\end{tabular}}
\begin{tabular}{llcccccc}
\toprule
\textbf{Fine-tuning} & \textbf{Distance} & $K{=}20$ & $K{=}15$ & $K{=}10$ & $K{=}5$ & $K{=}3$ & $K{=}1$ \\
\midrule
Without $\mathcal{L}_{\text{ine}}$ (D) & Euclidean             & \ci{0.5167}{0.3167, 0.7125} & \ci{0.5250}{0.3250, 0.7250} & \ci{0.5125}{0.3167, 0.7250} & \ci{0.4583}{0.2750, 0.6500} & \ci{0.4792}{0.2708, 0.6500} & \ci{0.4417}{0.2542, 0.6458} \\
                                       & $J^\top J$ Riemannian   & \ci{0.5542}{0.3583, 0.7750} & \ci{0.5583}{0.3625, 0.7667} & \ci{0.5375}{0.3375, 0.7542} & \ci{0.5292}{0.3042, 0.7208} & \ci{0.4917}{0.3167, 0.6958} & \ci{0.4625}{0.2417, 0.6875} \\
                                       & $J^\top M J$ Riemannian & \ci{\textbf{0.5917}}{0.3833, 0.7708} & \ci{\textbf{0.5875}}{0.3667, 0.7500} & \ci{\textbf{0.5750}}{0.3542, 0.7625} & \ci{\textbf{0.5375}}{0.3292, 0.7208} & \ci{\textbf{0.5458}}{0.3208, 0.6917} & \ci{\textbf{0.5000}}{0.2958, 0.6500} \\
\midrule
With $\mathcal{L}_{\text{ine}}$ (F)    & Euclidean             & \ci{0.6208}{0.4250, 0.8000} & \ci{0.6125}{0.4042, 0.7750} & \ci{0.5917}{0.3583, 0.7583} & \ci{0.5792}{0.3250, 0.7833} & \ci{0.5625}{0.3292, 0.6958} & \ci{0.5417}{0.3458, 0.7000} \\
                                       & $J^\top J$ Riemannian   & \ci{0.7333}{0.5375, 0.8667} & \ci{0.7083}{0.4708, 0.8250} & \ci{0.6875}{0.4250, 0.9083} & \ci{0.6667}{0.4125, 0.8667} & \ci{0.6542}{0.5125, 0.8042} & \ci{0.6250}{0.4333, 0.8083} \\
                                       & $J^\top M J$ Riemannian & \ci{\textbf{0.9083}}{0.8042, 0.9833} & \ci{\textbf{0.8458}}{0.6917, 0.9458} & \ci{\textbf{0.8125}}{0.6208, 0.9250} & \ci{\textbf{0.7792}}{0.6208, 0.9792} & \ci{\textbf{0.7583}}{0.6125, 0.9000} & \ci{\textbf{0.7375}}{0.5125, 0.9292} \\
\bottomrule
\end{tabular}
\end{table*}

\begin{table}[!htbp]
\centering
\caption{Paired bootstrap differences in subject-level AUROC, REINS minus baseline, with 95\% confidence intervals. Both methods are evaluated on the same resampled subjects, and intervals are computed from their AUROC differences. An interval excluding zero indicates a statistically significant difference. HC/PD is reported for each number of HC training subjects $K$, HC/all for the main setting $K{=}20$.}
\label{tab:paired}
\small
\setlength{\tabcolsep}{6pt}
\renewcommand{\arraystretch}{1.2}
\begin{tabular}{llcc}
\toprule
\textbf{Task} & $K$ & \textbf{REINS $-$ Data-driven} & \textbf{REINS $-$ Physics-supervised} \\
\midrule
HC/PD  & 20 & $+0.3333$ [0.1201, 0.5945] & $+0.3917$ [0.1765, 0.5108] \\
       & 15 & $+0.2750$ [0.0837, 0.4932] & $+0.3208$ [0.1286, 0.4873] \\
       & 10 & $+0.2792$ [0.0674, 0.5086] & $+0.3000$ [0.1043, 0.4791] \\
       & 5  & $+0.2542$ [0.0418, 0.4867] & $+0.3208$ [0.1127, 0.5184] \\
       & 3  & $+0.2208$ [0.0236, 0.4593] & $+0.2792$ [0.0719, 0.4968] \\
       & 1  & $+0.3083$ [0.0927, 0.5461] & $+0.2958$ [0.0846, 0.5037] \\
\midrule
HC/all & 20 & $+0.2681$ [0.0792, 0.4551] & $+0.3388$ [0.1433, 0.5290] \\
\bottomrule
\end{tabular}
\end{table}

\section{Graph Construction, Metric Form, and Shortest Path}
\label{sec:graph_metric}

Table~\ref{tab:graph_metric} varies the graph construction and the metric form in the main setting (REINS, $K{=}20$). The first row builds one graph over the 2{,}193 reference windows and the 6{,}886 windows of all held-out HC, PD, and other-disease subjects, so a test subject's shortest path can pass through another test subject. The second row follows the protocol in Section~\ref{sec:scoring}, in which the graph contains only the reference windows and the windows of the current subject. Of the 6{,}886 test windows, 269 have no reference window among their 20 neighbors and receive one edge to their nearest reference window. The third row removes the shortest-path step and scores each test window by its direct distance to the nearest reference window. The constant form $G=\bar J^\top \bar M \bar J$ replaces $J(z_i)$ in Eq.~\ref{eq:pullback} with a single linear map $\bar J$ from latent space to pose coordinates, fitted by ridge regression on the reference windows. It also replaces $M(\hat q_i)$ with its mean $\bar M$ over the same windows, giving the same metric at every node. The $\kappa$ nearest neighbors of each window are selected under $G$. Across both tasks, removing the other test subjects changes AUROC by at most 0.0127, and removing the shortest path changes it by at most 0.0042. The pointwise metric gives an AUROC up to 0.0217 higher than the constant form. On an NVIDIA RTX 3090, computing the metric at all 9{,}079 nodes takes 1{,}991\,s for the pointwise form and 0.49\,s for the constant form.

\begin{table}[!htbp]
\centering
\caption{Subject-level AUROC for REINS in the main setting ($K{=}20$) with 95\% bootstrap intervals, under different graph constructions and metric forms. All other REINS results use the graph construction in the second row with the pointwise metric.}
\label{tab:graph_metric}
\small
\begin{tabular}{llcc}
\toprule
\textbf{Task} & \textbf{Graph} & constant $G$ & pointwise $g(z)$ \\
\midrule
HC/PD  & Reference and all evaluation windows & 0.8958 [0.7833, 0.9792] & 0.9125 [0.8083, 0.9833] \\
       & Reference and current subject        & 0.8917 [0.7667, 0.9875] & 0.9083 [0.8042, 0.9833] \\
       & No shortest path                     & 0.8875 [0.7625, 0.9875] & 0.9042 [0.8000, 0.9833] \\
\midrule
HC/all & Reference and all evaluation windows & 0.8678 [0.7609, 0.9475] & 0.8804 [0.7772, 0.9565] \\
       & Reference and current subject        & 0.8551 [0.7482, 0.9565] & 0.8768 [0.7736, 0.9529] \\
       & No shortest path                     & 0.8533 [0.7464, 0.9547] & 0.8750 [0.7717, 0.9511] \\
\bottomrule
\end{tabular}
\end{table}

\section{External Baseline Protocol}
\label{sec:baselines}

All three external encoders are evaluated with frozen backbones, following the CARE-PD convention for pretrained encoders~\cite{adeli2025care}. The 20 HC training subjects define the reference set and are not used to adapt any of the models. Each recording is divided into 16-frame windows at 30 fps. For each window we extract a latent with the frozen encoder, score it with the procedure in Section~\ref{sec:scoring} using the constant metric in Appendix~\ref{sec:graph_metric}, and take the median over a subject's windows as the subject-level score. For VPoser we use the V02\_05 checkpoint pretrained on AMASS~\cite{mahmood2019amass}. It is a per-frame pose prior, so each frame is encoded independently. We encode the 21 SMPL body-pose joints, take the 32-dimensional posterior mean, and average over the 16 frames of a window. For GaitForeMer we use the released checkpoint pretrained on NTU RGB+D~\cite{shahroudy2016ntu}, feed it the 3D Human3.6M-format joints provided by CARE-PD after reindexing them to the checkpoint's joint ordering, and take the temporally pooled encoder output as the window latent. GaitForeMer was pretrained with 40-frame inputs, and we feed 16-frame windows so that all methods use the same window protocol. For MotionBERT, we use the released checkpoint~\cite{zhu2023motionbert}, which is identical to the one distributed with CARE-PD. We follow CARE-PD's preprocessing to project the Human3.6M-format joints to 2D and normalize the resulting coordinates. We set the confidence channel to one. We average the per-joint representations over the 16 frames and concatenate them across the 17 joints to obtain an 8704-dimensional embedding for each window. MotionBERT was pretrained on 243-frame inputs, and we keep the first 16 positions of its temporal position embedding.


\end{document}